\documentclass[sigconf]{acmart}

\usepackage{microtype}
\usepackage{graphicx}
\usepackage{subfigure}
\usepackage{amssymb}
\usepackage{amsmath}
\usepackage{commath}

\usepackage{paralist}
\usepackage{geometry}
\usepackage{natbib}
\usepackage{booktabs} % for professional tables
\usepackage[margin=1cm]{caption}
\DeclareMathOperator*{\concat}{\scalebox{1}[1.5]{$\parallel$}}

\usepackage{hyperref}

\usepackage{blindtext}

\author[Andreea Deac, Yu-Hsiang Huang, Petar Veli\v{c}kovi\'{c}, Pietro Li\`o, Jian Tang]{Andreea Deac$^{1,2}$,  Yu-Hsiang Huang$^2$, Petar Veli\v{c}kovi\'{c}$^1$, Pietro Li\`o$^1$, Jian Tang$^{2,3}$ \\[1ex]
\large{ $^1$ Department of Computer Science and Technology, University of Cambridge \\ $^2$ Mila -- Qu\'ebec Artificial Intelligence Institute \quad $^3$ HEC Montr\'eal}
}

\begin{document}

\title[\bf Drug-Drug Adverse Effect Prediction with Graph Co-Attention]{{\bf Drug-Drug Adverse Effect Prediction\\ with Graph Co-Attention}}

\begin{abstract}
Complex or co-existing diseases are commonly treated using drug combinations, which can lead to higher risk of adverse side effects. The detection of polypharmacy side effects is usually done in Phase IV clinical trials, but there are still plenty which remain undiscovered when the drugs are put on the market. Such accidents have been affecting an increasing proportion of the population (15\% in the US now) and it is thus of high interest to be able to predict the potential side effects as early as possible. Systematic combinatorial screening of possible drug-drug interactions (DDI) is challenging and expensive. However, the recent significant increases in data availability from pharmaceutical research and development efforts offer a novel paradigm for recovering relevant insights for DDI prediction. Accordingly, several recent approaches focus on curating massive DDI datasets (with millions of examples) and training machine learning models on them. Here we propose a neural network architecture able to set state-of-the-art results on this task---using the type of the side-effect and the molecular structure of the drugs alone---by leveraging a co-attentional mechanism. In particular, we show the importance of integrating joint information from the drug pairs early on when learning each drug's representation.
\end{abstract}

\maketitle

\section{Introduction}
Diseases are often caused by complex biological processes which cannot be treated by individual drugs and thus introduce the need for concurrent use of multiple medications. Similarly, drug combinations are needed when patients suffer from multiple medical conditions. However, the downside of such treatment (commonly referred to as \emph{polypharmacy}) is that it increases the risk of adverse side effects, caused by the chemical-physical incompatibility of the drugs. Such drug-drug interactions (DDIs) have become a serious issue in recent times, affecting nearly 15\% of the US population \cite{Kantor}, with treatment costs exceeding \$177 billion/year \cite{Ernst}.\\ \\ 
Detection of polypharmacy side effects remains a challenging task to this day: systematic combinatorial screening is expensive and, while the search for such side effects is usually done during clinical testing, the small scale of the trials and the rarity of adverse DDI means that many adverse drug reactions (ADRs) are unknown when the drugs reach the market. Moreover, the immense cost of designing a drug makes it desirable to be able to predict ADRs as early as possible in the pipeline. \\ \\ 
In this work, we propose the use of deep learning to ameliorate this problem. One of deep learning's main advantages is the ability to perform automated feature extraction from raw data. This means that, provided the existence of sufficiently large datasets, models can now be developed without depending on manually engineered features, which are expensive to obtain and require expert domain knowledge \cite{Goodfellow}. In this work, we leverage such a large dataset \cite{TWOSIDES}, consisting of $\sim$4.5M examples of \emph{(drug, drug, interaction)} tuples, to demonstrate that these models are indeed viable in the DDI domain.\\ \\
By looking at the chemical structures of the drug pairs, our model learns drug representations which incorporate \textit{joint drug-drug information}. We then predict whether a drug-drug pair would cause side effects and what their types would be. Our method outperforms previous state-of-the-art models in terms of predictive power, while using just the molecular structure of the drugs. It thus allows for the method to be applied to novel drug combinations, as well as permitting the detection to be performed at the preclinical phase rather than in the drug discovery phase, leveraging clinical records.\\ \\
Our approach paves the way to a new and promising direction in analysing polypharmacy side effects with machine learning, that is in principle applicable to any kind of interaction discovery task between \emph{structured} inputs (such as molecules), especially under availability of large quantities of labelled examples.

\begin{figure*}
    \includegraphics[width=0.48\linewidth]{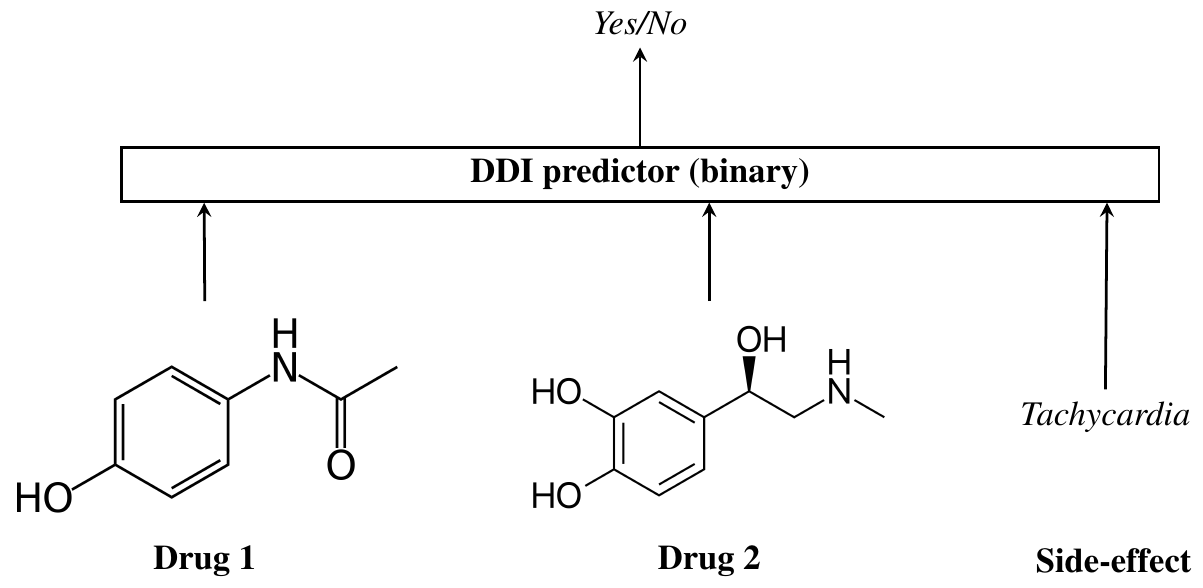} \hfill
    \includegraphics[width=0.48\linewidth]{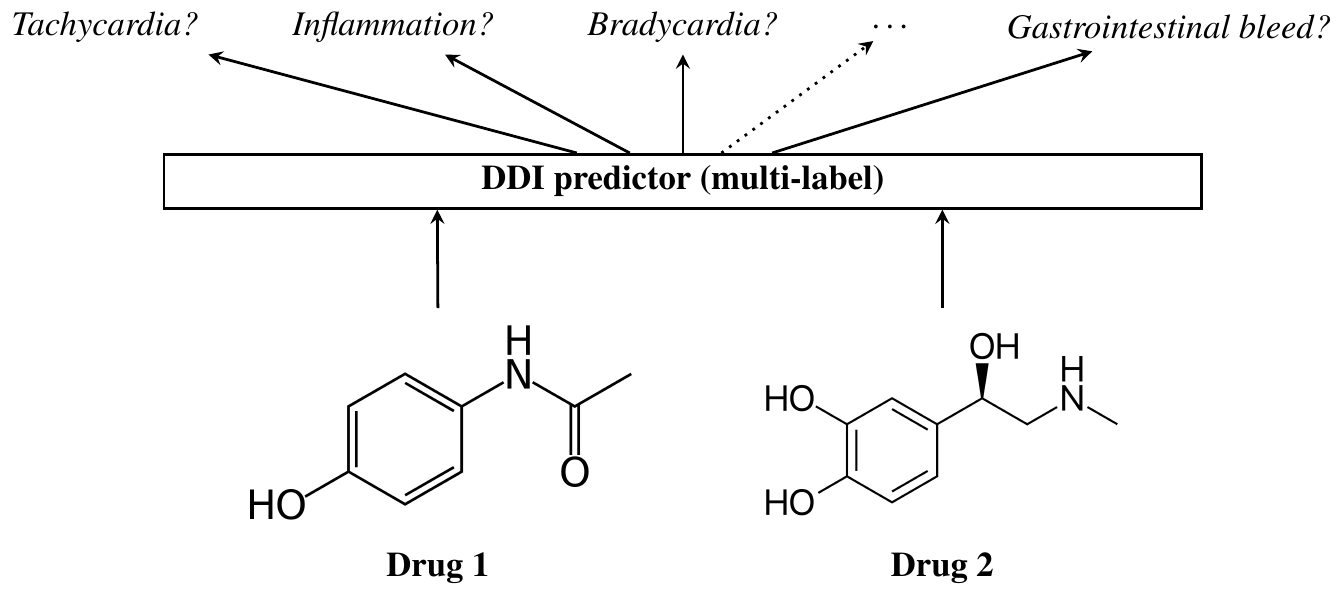}
    \caption{An overview of the binary (\emph{left}) and multi-label (\emph{right}) drug-drug interaction (DDI) task. In both cases, two drugs (represented as their molecular structures) are provided to the model in order to predict existence or absence of adverse interactions. For binary classification, the DDI predictor is also given a particular side effect as input, and is required to specifically predict existence or absence of it. For multi-label classification, the DDI predictor simultaneously predicts existence or absence of all side-effects under consideration.}
    \label{fig:setup}
\end{figure*}

\section{Related work}
Our work finds itself at the intersection of two domains: computational methods for prediction of side effects caused by DDI and neural networks for graph-structured data. As such, a review of related advances in each area will be presented here.\\ \\ 
As manually examining drug combinations and their possible side effects cannot be done exhaustively, computational methods were first developed to identify the drug pairs which create a response higher than the additive response they would cause if they did not interact \cite{Ryall}. This was previously done by framing the task as a binary classification problem and designing machine learning models (na\"{i}ve Bayes, logistic regression, support vector machines) which predict the probability of a DDI, using the measurement of cell viability \cite{Huang2014, Sun, Zitnik2016, Chen, Shi}. Other related approaches considered dose-effect curves \cite{Bansal, Takeda} or synergy and antagonism \cite{Lewis}. \\ \\
An alternative way of approaching the task is provided by models which use the assumption that drugs with similar features are more likely to interact \cite{Gottlieb, Vilar2012, Huang2014, Li2015, Zitnik2015, Sun, Li2017}. Using features such as the chemicals' structures, individual drug side effects and interaction profile fingerprints, the models use unsupervised or semi-supervised techniques (clustering, label propagation) in order to find DDIs. Alternatively, restricted Boltzmann machines and matrix factorization were used to combine different types of similarities by learning latent representations \cite{Cao2015, Wang2013}.\\ \\
However, all these methods are limited to either providing the likelihood of a DDI (but not its type if one exists), or lack applicability in inductive settings. Decagon \cite{Decagon} and the Multitask Dyadic Prediction in \cite{Dyadic} are two methods which overcome these challenges, and are thus going to be used as baselines against which our work will be compared. Multitask Dyadic Prediction is a proximal gradient method which uses substructure fingerprints to construct the drug feature representations. Similarly to our work, it has access only to the chemical structure of the drug.\\ \\
Decagon, on the other hand, improves predictive power further by including additional relational information with \emph{protein} targets of interest. Specifically, Decagon leverages this information by applying a \emph{graph convolutional neural network} architecture over a graph corresponding to the interactions between pairs of drugs, pairs of proteins and drug-protein pairs, treating discovery of novel DDIs as a \emph{link prediction} task in the graph. While the protein-related auxiliary information is highly beneficial for the algorithm to use, it could also be expensive to obtain. \\ \\ 
Compared to previous methods, our contribution is a model which learns a robust representation of drugs by leveraging joint information early on in the learning process. This allows it to bring an improvement in terms of predictive power, while maintaining an inductive setup where the model indicates the types of the possible side effects by just looking at chemical structure of the drugs.\\ \\
Our model builds up on a large existing body of work in \emph{graph convolutional networks} \cite{bruna2013spectral,defferrard2016convolutional,kipf2016semi,MPNN,GAT}, that have substantially advanced the state-of-the-art in many tasks requiring graph-structured input processing (such as the chemical representation \cite{MPNN,de2018molgan,you2018graph} of the drugs leveraged here). Furthermore, we build up on work proposing \emph{co-attention} \cite{lu2016hierarchical,Xmod} as a mechanism to allow for \emph{individual set-structured datasets} (such as nodes in multimodal graphs) to interact. Overall, these (and related) techniques correspond to one of the latest major challenges of machine learning \cite{bronstein2017geometric,hamilton2017representation,battaglia2018relational}, with transformative potential across a wide spectrum of potential applications (not only limited to the biochemical domain).

\begin{figure*}
    \centering
    \includegraphics[width=\linewidth]{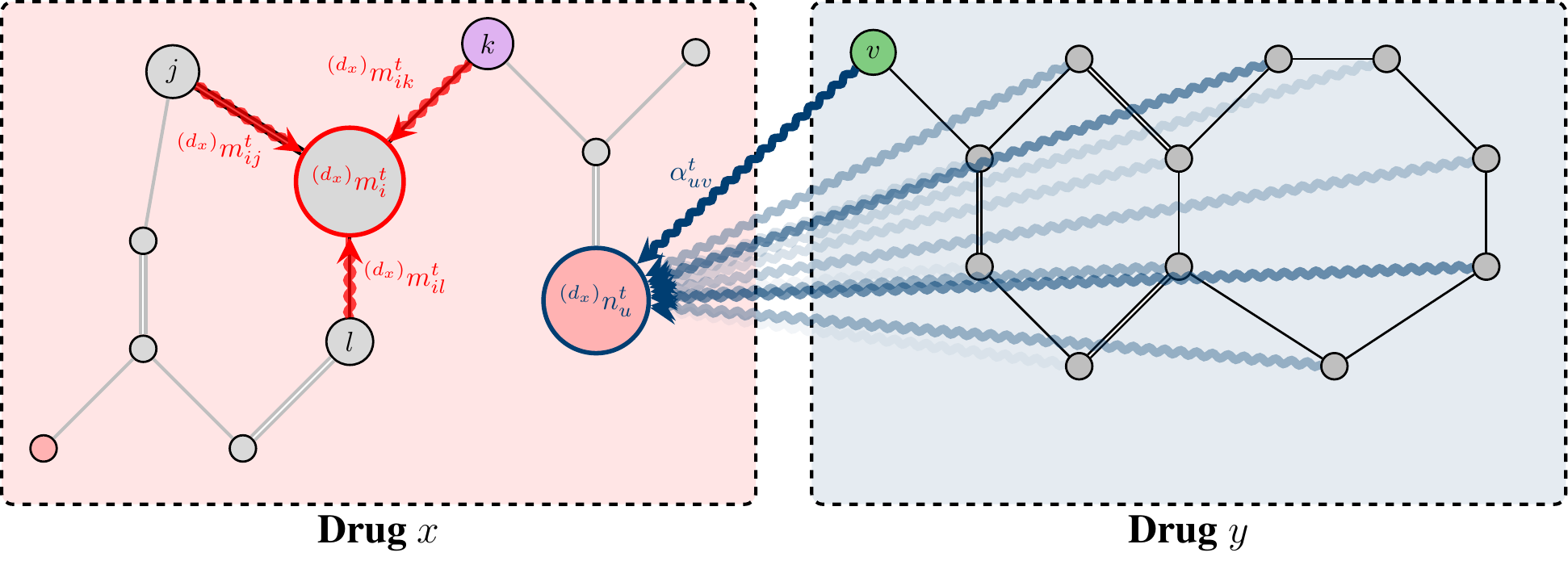}
    \caption{The illustration of a single step of message passing (computing the \emph{inner message}, $^{(d_x)}m_i^t$), and co-attention (computing the \emph{outer message} $^{(d_x)}n_u^t$) on two nodes ($i$ and $u$) of drug $x$.}
    \label{fig:mpass}
\end{figure*}

\section{Architecture}
In this section, we will present the main building blocks used within our architecture for drug-drug interaction prediction. This will span a discussion of the way the input to the model is encoded, followed by an overview of the individual computational steps of the model. Lastly, we will specify the loss functions optimised by the models.
\label{bin_methods}
\subsection{Inputs}
The \emph{drugs}, $d_x$, are represented as \emph{graphs} consisting of atoms, $a^{(d_x)}_i$ as nodes, and \emph{bonds} between those atoms $\left(a^{(d_x)}_i, a^{(d_x)}_j\right)$ as edges. For each atom, the following input features are recorded: the \emph{atom number}, the number of \emph{hydrogen atoms} attached to this atom, and the \emph{atomic charge}. For each bond, a discrete \emph{bond type} (e.g. single, double etc.) is encoded as a learnable input edge vector, $e_{ij}^{(d_x)}$. The \emph{side effects}, $se_z$, are one-hot encoded from a set of 964 side effects.\\ \\
The input to our model varies depending on whether we are performing \emph{binary classification} for a given side effect, or \emph{multi-label classification} for \emph{all} side effects at once:
\begin{itemize}
    \item For binary classification (Figure \ref{fig:setup} (Left)), the input to our model is a \emph{triplet} of two drugs and a side effect $(d_x, d_y, se_z)$, requiring a binary decision on whether drugs $x$ and $y$ adversely interact to cause side effect $z$.
    \item For multi-label classification (Figure \ref{fig:setup} (Right)), the input to our model is a pair of two drugs $(d_x, d_y)$, requiring 964 \emph{simultaneous} binary decisions on whether drugs $x$ and $y$ adversely interact to cause \emph{each} of the considered side effects. Note that, in terms of learning pressure, this model requires more robust \emph{joint representations} of pairs of drugs---as they need to be useful for all side-effect predictions \emph{at once}.
\end{itemize}
In both cases, the model returns a \emph{score} associated with the likelihood of a particular side effect occurring (higher scores implying larger likelihoods). This score can then be appropriately \emph{thresholded} to obtain a viable classifier.

\subsection{Message passing}
Within each of the two drugs separately, our model applies a series of \emph{message passing} \citep{MPNN} layers. Herein, nodes are allowed to send arbitrary \emph{vector messages} to each other along the edges of the graph, and each node then aggregates all the messages sent to it.\\ \\
Let $^{(d_x)}h_i^{t}$ denote the \emph{features} of atom $i$ of drug $x$, at time step $t$. Specially, initially these are set to projected input features, i.e.: \begin{equation}^{(d_x)}h_i^0 = f_{i}\left(a_i^{(d_x)}\right)
\end{equation}
where $f_{i}$ is a small multilayer perceptron (MLP) neural network. In all our experiments, this MLP (and all subsequently referenced MLPs and projections) projects its input to 32 features.\\ \\
Considering atoms $i$ and $j$ of drug $x$, connected by a bond with edge vector $e_{ij}^{(d_x)}$, we start by computing the \emph{message}, $^{(d_x)}m_{ij}^t$, sent along the edge $j \rightarrow i$. The message takes into account both the features of node $j$ and the features of the edge between $i$ and $j$: 
\begin{equation}\label{eqn:fst}
    ^{(d_x)}m_{ij}^{t} = f_e^t\left(e_{ij}^{(d_x)}\right) \odot f_v^t\left({}^{(d_x)}h_j^{t-1}\right)
\end{equation}
where $f_e^t$ and $f_v^t$ are small MLPs, and $\odot$ is elementwise vector-vector multiplication.\\ \\
Afterwards, every atom $i$ \emph{aggregates} the messages sent to it via summation. This results in an \emph{internal message} for atom $i$, $^{(d_x)}m_i^t$: 
\begin{equation}
    ^{(d_x)}m_i^t = \sum_{\forall j \in N(i)} {}^{(d_x)}m_{ij}^t
\end{equation}
where the neighbourhood, $N(i)$, defines the set of atoms linked to $i$ by an edge.
\subsection{Co-attention}
Message passing provides a robust mechanism for encoding within-drug representations of atoms. For learning an appropriate \emph{joint} drug-drug representation, however, we allow atoms to interact \emph{across} drug boundaries via a \emph{co-attentional mechanism} \citep{Xmod}.\\ \\
Consider two atoms, $i$ and $j$, of drugs $x$ and $y$, respectively. Let their features at time step $t$ be $^{(d_x)}h_i^{t}$ and $^{(d_y)}h_j^{t}$, just as before. For every such pair, we compute an \emph{attentional coefficient}, $\alpha_{ij}^t$ using a simplified version of the Transformer \cite{Transformer} attention mechanism:
\begin{equation} \label{asim}
    \alpha_{ij}^t = \mathrm{softmax}_j\left(\left<{\bf W}_k^t {}^{(d_x)}h_i^{t-1},  {\bf W}_k^t {}^{(d_y)}h_j^{t-1}\right>\right)
\end{equation}
where ${\bf W}_k^t$ is a learnable \emph{projection matrix}, $\left<\cdot, \cdot\right>$ is the \emph{inner product}, and the softmax is taken across \emph{all} nodes $j$ from the second drug. The coefficients $\alpha_{ij}^t$ may be interpreted as the \emph{importance} of atom $j$'s features to atom $i$.\\ \\
These coefficients are then used to compute an \emph{outer message} for atom $i$ of drug $x$, $^{(d_x)}n_i^t$, expressed as a \emph{linear combination} (weighted by $\alpha_{ij}^t$) of all (projected) atom features from drug $y$:
\begin{equation} \label{attn}
    ^{(d_x)}n_i^t = \sum_{\forall j \in d_y} \alpha_{ij}^t\cdot {\bf W}_v^t{}^{(d_y)}h_j^{t-1}
\end{equation}
where ${\bf W}_v^t$ is a learnable projection matrix.\\ \\
It was shown recently that \emph{multi-head attention} can stabilise the learning process, as well as learn information at different conceptual levels \citep{GAT,DeepInf}. As such, the mechanism of Equation \ref{attn} is independently replicated across $K$ attention \emph{heads} (we chose $K=8$), and the resulting vectors are \emph{concatenated} and \emph{MLP-transformed} to provide the final outer message for each atom. %Additionally, a leaky ReLU \citep{LeakyReLU} nonlinearity is applied to the result:
\begin{equation} \label{mhatt}
    ^{(d_x)}n_i^t = f_o^t\left(\concat_{k=1}^K \sum_{\forall j \in d_y} {}^{(k)}\alpha_{ij}^t\cdot {}^{(k)}{\bf W}_v^t{}^{(d_y)}h_j^{t-1}\right)
\end{equation}
where $f_o^t$ is a small MLP and $\|$ is featurewise vector concatenation.\\ \\
The computation of Equations \ref{asim}--\ref{mhatt} is replicated analogously for outer messages from drug $x$ to atoms of drug $y$.\\ \\
We have also attempted to use other popular attention mechanisms for computing the $\alpha_{ij}^t$ values---such as the original Transformer \cite{Transformer}, GAT-like \cite{GAT} and tanh \cite{bahdanau2014neural} attention---finding them to yield weaker performance than the approach outlined here.\\ \\
A single step of message passing and co-attention, as used by our model, is illustrated by Figure \ref{fig:mpass}.

\begin{figure}
    \centering
    \includegraphics[width=\linewidth]{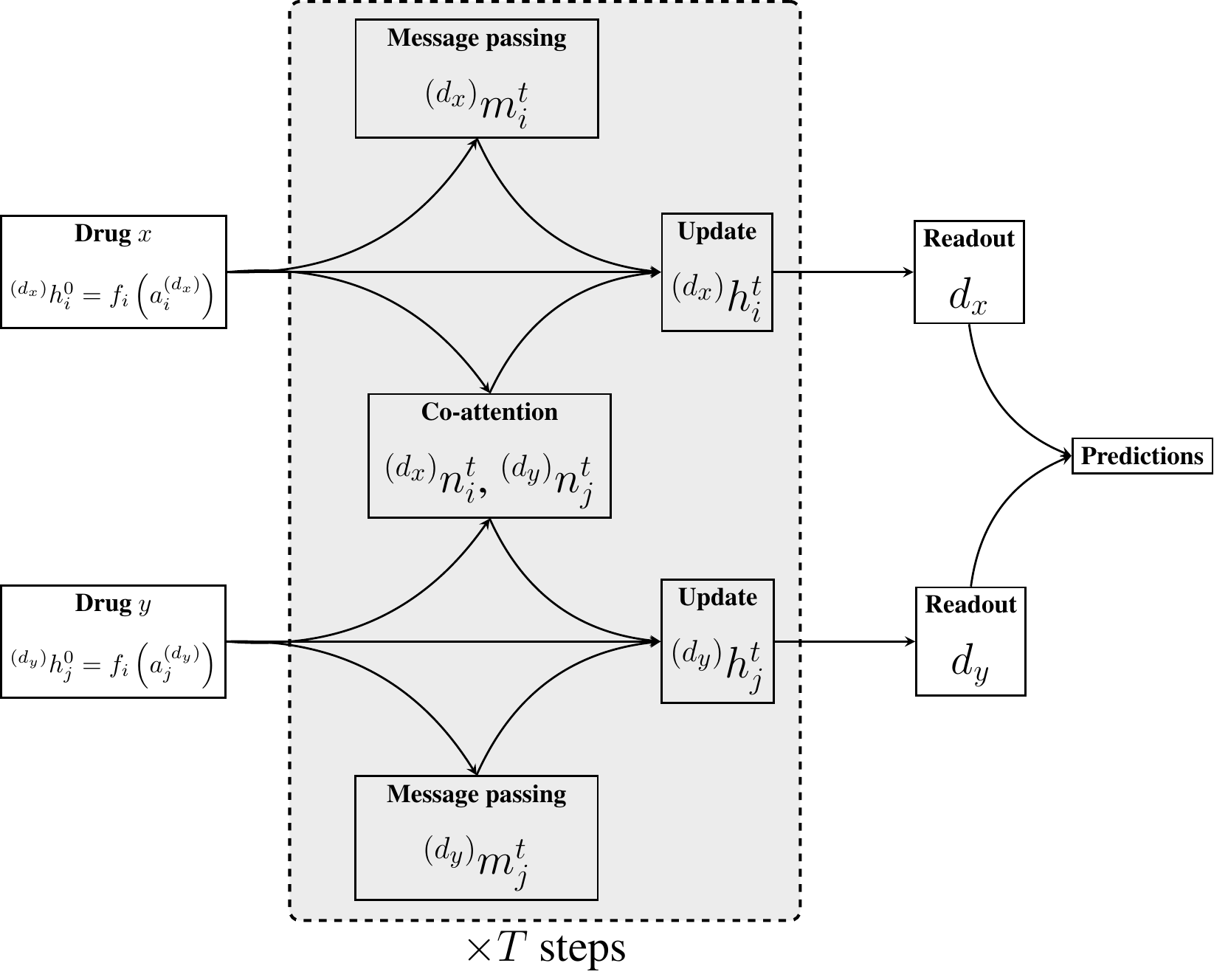}
    \caption{A high-level overview of our DDI model. The next-level features of atom $i$ of drug $x$, ${}^{(d_x)}h_i^t$, are derived by combining its \emph{input features}, ${}^{(d_x)}h_i^{t-1}$, its \emph{inner message}, ${}^{(d_x)}m_i^t$, computed using message passing, and its \emph{outer message}, ${}^{(d_x)}n_i^t$, computed using co-attention over the second drug, $d_y$.}
    \label{fig:caddi}
\end{figure}

\subsection{Update function}
Once the inner messages, $^{(d_x)}m_i^t$ (obtained through message passing), as well as the outer messages, $^{(d_x)}n_i^t$ (obtained through co-attention) are computed for every atom $i$ of each of the two drugs ($d_x$/$d_y)$, we use them to derive the next-level features, $^{(d_x)}h_i^t$, for each atom.\\ \\
At each step, this is done by aggregating (via summation) the previous features (representing a \emph{skip connection} \citep{ResNet}), the inner messages and outer messages, followed by \emph{layer normalisation} \citep{LayerNorm}:
\begin{equation}\label{eqn:upd}
    ^{(d_x)}h_i^t = \mathrm{LayerNorm}\left({}^{(d_x)}h_i^{t-1} + {}^{(d_x)}m_i^t + {}^{(d_x)}n_i^t\right)
\end{equation}
The operations of Equations \ref{eqn:fst}--\ref{eqn:upd} are then repeated for $T$ propagation steps---here, we set $T=3$. Refer to Figure \ref{fig:caddi} for a complete visualisation of our architecture.\\ \\
As will be demonstrated in the Results section, using co-attention to enable the model to propagate the information between two drugs from the beginning---thus learning a \emph{joint} representation of the drugs---valuably contributes to the predictive power of the model.

\subsection{Readout and Scoring}

%TODO: Reference
%TODO: Think whether to describe all transH, transE, transR
As we're making predictions on the level of entire drug-drug pairs, we need to compress individual atom representations into drug-level representations. In order to obtain the drug-level vectors $d_x$, we apply a simple \emph{summation} of its constituent atom feature vectors after the final layer (i.e. after $T$ propagation steps have been applied): 
\begin{equation}
    d_x = \sum_{\forall j \in d_x} f_r\left({}^{(d_x)}h_j^T\right)
\end{equation}
where $f_r$ is a small MLP.\\ \\
Once the drug vectors are computed for both $d_x$ and $d_y$, these can be used to make predictions on side effects---and the exact setup varies depending on the type of classification.
\subsubsection{Binary classification}
In the binary classification case, the side effect vector $se_z$ is provided as input. We then leverage a scoring function $f$ similar to the one used by Yoon \emph{et al.} \cite{Scoring} to express the likelihood of this side effect occuring: 
\begin{equation}
    %loss_1 = \norm{M_h d_1 + SE - M_t d_2}^2_2 \qquad loss_2 = \norm{M_h d_2 + SE - M_t d_1}^2_2
    f(d_x, d_y, se_z) = \norm{{\bf M}_h d_x + se_z - {\bf M}_t d_y}^2_2 + \norm{{\bf M}_h d_y + se_z - {\bf M}_t d_x}^2_2
\end{equation}
%\begin{equation}
%    loss_{total} = loss_1 + loss_2
%\end{equation}
where $\|\cdot\|_2$ is the $L_2$ norm, and ${\bf M}_h$ and ${\bf M}_t$ represent the \emph{head node} and \emph{tail node} space mapping matrices, respectively.\\ \\
The model is then trained end-to-end with gradient descent to optimise a \emph{margin-based} ranking loss: 
\begin{equation}
    \mathcal{L}= \sum_{d_x, d_y, se_z} \sum_{\Tilde{d_x}, \Tilde{d_y}, se_z} \max(0, \gamma - f(d_x, d_y, se_z) - f(\Tilde{d_x}, \Tilde{d_y}, se_z)))
\end{equation}
where $(d_x, d_y)$ is a drug-drug pair exhibiting side effect $se_z$, and $(\Tilde{d_x}, \Tilde{d_y})$ is a drug-drug pair not exhibiting it. $\gamma > 0$ is the \emph{margin} hyperparameter.

\subsubsection{Multi-label classification}
% NB I needed to manually split parametrised here...
Here, all side effects are predicted simultaneously, and accordingly we define a \emph{prediction layer}, param-etrised by a learnable weight matrix ${\bf W}_p$ and bias $b_p$. This layer consumes the concatenation of the two drug vectors and projects it into a score for each of the 964 side effects. These scores are then converted into probabilities, $y_{xy}^z$, of each side effect, $z$, occurring between drugs $x$ and $y$ using the \emph{logistic sigmoid} nonlinearity, $\sigma$, applied elementwise:
\begin{equation}
    y_{xy}^z = \sigma\left({\bf W}_p[d_x\|d_y] + b_p\right)_z
\end{equation}
We can then once again train the model end-to-end with gradient descent, using \emph{binary cross-entropy} against the ground truth value, $\hat{y}_{xy}^z$ (which is a binary label indicating whether side effect $z$ actually occurs between drugs $x$ and $y$). The loss function is as follows:
\begin{equation}
    \mathcal{L}_{BCE} = \sum_{d_x, d_y, z} \hat{y}_{xy}^z \log{y_{xy}^z} + \left(1 - \hat{y}_{xy}^z\right) \log{\left(1 - y_{xy}^z\right)}
\end{equation}

\section{Dataset and Preprocessing}
\label{bin_data}
The drug-drug interaction data was chosen to be in agreement with the dataset used by Decagon \cite{Decagon}. It is obtained through filtering the TWOSIDES side-effect dataset \cite{TWOSIDES}, which consists of associations which cannot be clearly attributed to either drug alone (that is, those side-effects reported for any individual drug in the OFFSIDE dataset \cite{TWOSIDES}) and it comprises 964 polypharmacy side-effect types that occurred in at least 500 drug pairs. \\ \\
The data format used within the dataset is:
\begin{inparaenum}
    \item The compound ID of drug 1;
    \item The compound ID of drug 2;
    \item Side effect concept identifier;
    \item Side effect name.
\end{inparaenum}
In order to obtain the drug-related features, we look up the compound IDs in the PubChem database. This provides us with the molecular structure of each drug:
\begin{itemize}
    \item Atoms, including type, charge and coordinates for each atom;
    \item Bonds, including type and style for each bond;
    \item SMILES string.
\end{itemize}
In order to compensate for the fact that TWOSIDES contains only positive samples, appropriate \emph{negative sampling} is performed for the binary classification task:
\begin{itemize}
    \item During training, tuples $(\Tilde{d_x}, \Tilde{d_y}, se_z)$, where $\Tilde{d_x}$ and $se_z$ are chosen from the dataset and $\Tilde{d_y}$ is chosen at random from the set of drugs different from $d_y$ in the true samples $(\Tilde{d_x}, d_y, se_z)$.
    \item During validation and testing, we simply randomly sample two distinct drugs which do not appear in the positive dataset.
\end{itemize}
The full dataset consists of 4,576,785 positive examples. As we aim to sample a balanced number of positive and negative examples, the overall dataset size for training our model is 9,153,570 examples.

\section{Experimental setup}

As previously described, our model consists of $T=3$ propagation blocks (featuring a message-passing layer followed by a co-attentional layer with $K=8$ attention heads). Each block computes 32 intermediate features. To further regularise the model, dropout \cite{Dropout} with $p=0.2$ is applied to the output of every intermediate layer. The model is initialised using Xavier initialisation \cite{Xavier} and trained on mini-batches of 200 drug-drug pairs, using the Adam SGD optimiser \cite{Adam} for 30 epochs. The learning rate after $t$ iterations, $\eta_t$, is derived using an exponentially decaying schedule:
\begin{equation}
    \eta_t = 0.001 \cdot 0.96^{t\cdot 10^{-6}}
\end{equation}
We will refer to this model as \textbf{MHCADDI} (\emph{multi-head co-attentive drug-drug interactions}), and \textbf{MHCADDI-ML} for multi-label classification. We compare our method with the following baselines:
\begin{itemize}
    \item The work of Jin \emph{et al.} \cite{Dyadic} (to be referred to as \emph{Drug-Fingerpri-nts}), where binary classification is performed through analys-ing the \emph{molecular fingerprints} of drugs.
    \item \emph{Decagon} \cite{Decagon}, where \emph{drug-protein interactions} and \emph{protein-protein interactions} are considered in as \emph{additional inputs} to the model, and the entire interaction graph is simultaneously processed by a graph convolutional network.
    \item We also report the results of all the baselines used by \emph{Decagon}, such as the RESCAL \cite{nickel2011three} and DEDICOM \cite{papalexakis2017tensors} tensor decomposition methods and DeepWalk \cite{perozzi2014deepwalk}.
\end{itemize}
In addition, we perform detailed ablation studies to stress the importance of various components of MHCADDI. Specifically, the following baselines are also evaluated:
\begin{itemize}
    \item An architecture (to be referred to as \emph{MPNN-Concat}) where the drug representations are learnt separately, solely using \emph{internal message passing}. The side effect probability is then computed by concatenating the individual drug representations. This serves to demonstrate the importance of jointly learning the drug embeddings (by e.g. co-attention).
    \item An architecture (to be referred to as \emph{Late-Outer}) where the outer messages are only considered at the end of the node updating procedure. Each propagation step $t$ of the MPNN function creates ``clusters'' with information from neighbouring atoms within the same drug at a distance of at most $t$ hops away.  The outer messages are then computed at different granularity levels and integrated at the end of the pipeline. Similarly as before, this serves to demonstrate the importance of simultaneously performing internal and external feature extraction.
    \item An architecture (to be referred to as \emph{CADDI}) where there is only $K=1$ attention head.
\end{itemize}
We omit comparisons with traditional machine learning baselines (such as support vector machines or random forests) as prior work \cite{Dyadic} has already found them to be significantly underperforming compared to neural network approaches on this task.

\begin{table}[t]
\centering
\caption{Comparative evaluation results after stratified 10-fold crossvalidation.}
  \begin{tabular}{l c c}
\toprule 
 & {\bf AUROC} \\ %& {\bf AUPRC} \\ 
 \midrule
  {\bf Drug-Fingerprints} \cite{Dyadic} & $0.744$ \\ \midrule %& $0.239$\\
 {\bf RESCAL} \cite{nickel2011three} & 0.693\\
 {\bf DEDICOM} \cite{papalexakis2017tensors} & 0.705\\
 {\bf DeepWalk} \cite{perozzi2014deepwalk} & 0.761\\
 {\bf Concatenated features} \cite{Decagon} & 0.793\\
 {\bf Decagon} \cite{Decagon} & $0.872$ \\ \midrule %& $0.832$ \\
 {\bf MHCADDI} (ours) & {$\bf 0.882$}  \\ %& $0.883$ \\ \midrule
 {\bf MHCADDI-ML} (ours) &  $0.819$  \\
 
\bottomrule
\end{tabular}
\label{table:roc_auc}
\end{table}

\begin{table}[t]
\centering
\caption{Ablation study for various aspects of the MHCADDI model.}
  \begin{tabular}{l c c}
\toprule 
 & {\bf AUROC} \\ %& {\bf AUPRC} \\ 
 \midrule
 {\bf MPNN-Concat}  & $0.661$ \\ %& $0.$\\
 {\bf Late-Outer}  & $0.724$ \\ %& $0.$\\
 {\bf CADDI} & $0.778$ \\ % & $0.$ \\\midrule
 {\bf MHCADDI} & {$\bf 0.882$}\\
\bottomrule
\end{tabular}
\label{table:roc2}
\end{table}

\begin{table}
\centering
\caption{Top-5 and bottom-5 side effects by AUROC values.}
  \begin{tabular}{l c}
  \multicolumn{2}{c}{\bf Highest} \\
\toprule 
 {\bf Side effect} & {\bf AUROC} \\ %& {\bf AUPRC} \\ 
 \midrule
 {\bf Tooth impacted}  & $0.903$ \\ %& $0.$\\
 {\bf Nasal polyp}  & $0.902$ \\ %& $0.$\\
 {\bf Cluster headache} & $0.900$ \\ % & $0.$ \\\midrule
 {\bf Balantis} & {$0.892$}\\
 {\bf Dysphemia} & {$0.889$}\\
\bottomrule
\end{tabular}
\newline
 \begin{tabular}{l c}
   \multicolumn{2}{c}{\bf Lowest} \\
\toprule 
 {\bf Side effect} & {\bf AUROC} \\ %& {\bf AUPRC} \\ 
 \midrule
 {\bf Neonatal respiratory distress syndrome}  & $0.670$ \\ %& $0.$\\
 {\bf Aspergillosis}  & $0.683$ \\ %& $0.$\\
 {\bf Obstructive uropathy} & $0.690$ \\ % & $0.$ \\\midrule
 {\bf Icterus} & {$0.712$}\\
 {\bf HIV disease} & {$0.712$}\\
\bottomrule
\end{tabular}
\label{table:roc4}
\end{table}

\section{Results}

\label{results}

\subsection{Quantitative Results}

We perform stratified 10-fold crossvalidation on the derived dataset to evaluate our models and compare them against previously published baselines. For each model, we report the area under the receiver operating characteristic (ROC) curve averaged across the 10 folds. The results of our study may be found in Tables \ref{table:roc_auc}--\ref{table:roc2}.\\ \\ 
From this study, we may conclude that learning \emph{joint} drug-drug representations by simultaneously combining internal message-passing layers and co-attention between drugs is a highly beneficial approach, consistently outperforming the strong baselines considered here (even when additional data sources are used, as in the case of Decagon). Furthermore, through observing the comparisons between several variants of our architecture, more detailed conclusions easily follow:
\begin{itemize}
    \item The fact that all our co-attentive architectures outperformed MPNN-Concat implies that it is beneficial to learn drug-drug representations \emph{jointly} rather than separately.
    \item Furthermore, the outperformance of Late-Outer by (MH)CA-DDI further demonstrates that it is useful to provide the cross-modal information \emph{earlier} in the learning pipeline.
    \item Lastly, the comparative evaluation of MHCADDI against CADDI further shows the benefit of inferring \emph{multiple mechanisms} of drug-drug interaction simultaneously, rather than anticipating only one.
\end{itemize}
Finally, we can observe that, despite the additional learning challenges present when doing multi-label classification (in the case of MHCADDI-ML), our jointly learned representations are still competitive with all other baselines. In particular, our method significantly outperforms Drug-Fingerprints---which also uses a multi-label objective.\\ \\
As an additional ablation study, similarly to the analysis performed by Decagon \cite{Decagon}, for one testing fold we stratified the predictions of our model across all side effects, computing side effect-level AUROC values. These values are given for five of the best- and worst- performing side effects in \ref{table:roc4}. There is a slight correpondence to the performance of Decagon---with Icterus appearing as one of their worst-performing side effects and a different polyp appearing as one of their best-performing ones. Overall, we do not find that our distribution of per-side-effect AUROC values deviates from the mean value any more substantially than what was reported for Decagon. This implies that our model does not drastically sacrifice performance on some side-effects---at least, not compared to the prior baseline methods which use auxiliary data sources.

\subsection{Qualitative Results}

\begin{figure}
    \centering
    \includegraphics[width=0.3\linewidth]{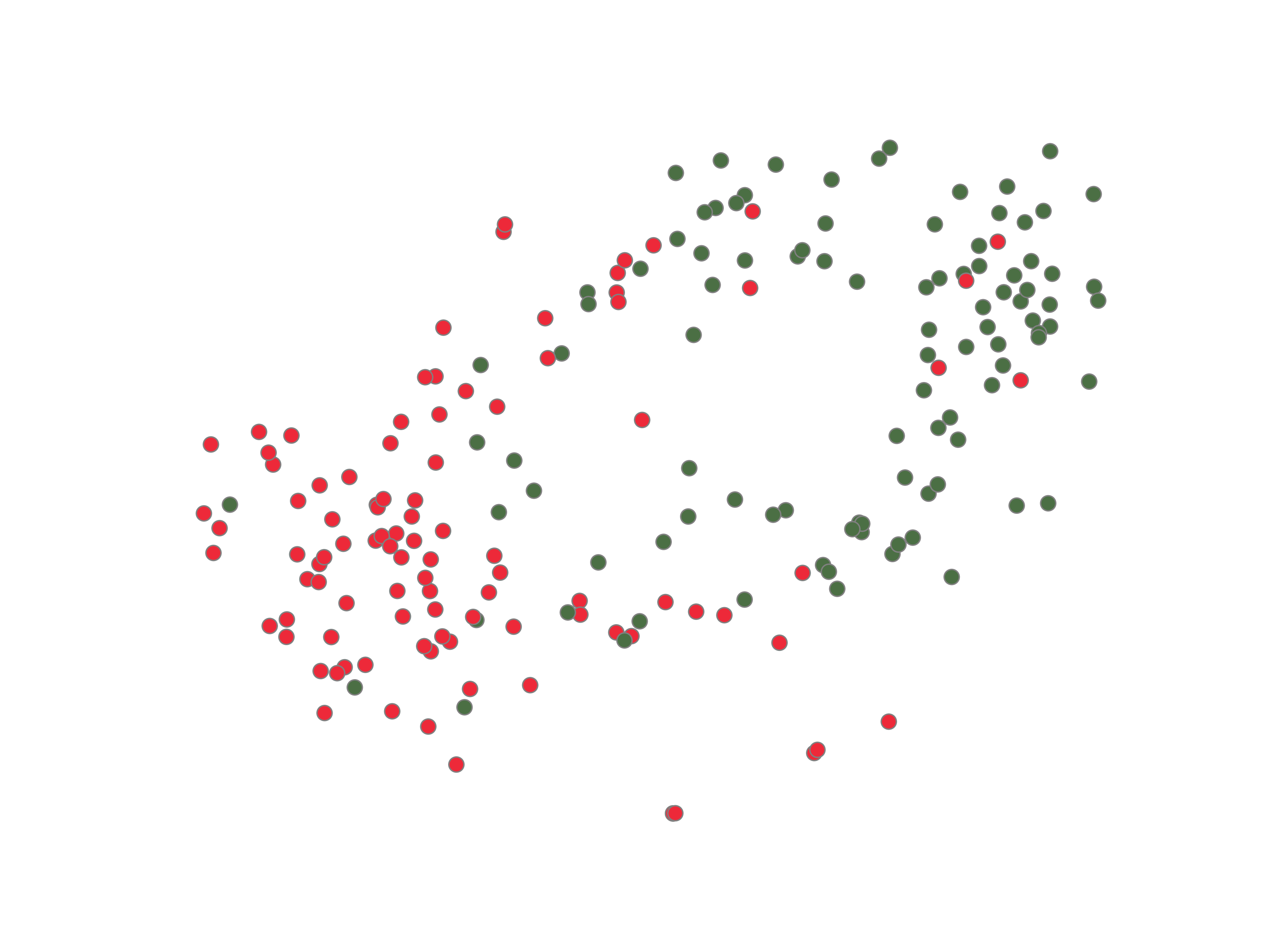}
    \includegraphics[width=0.3\linewidth]{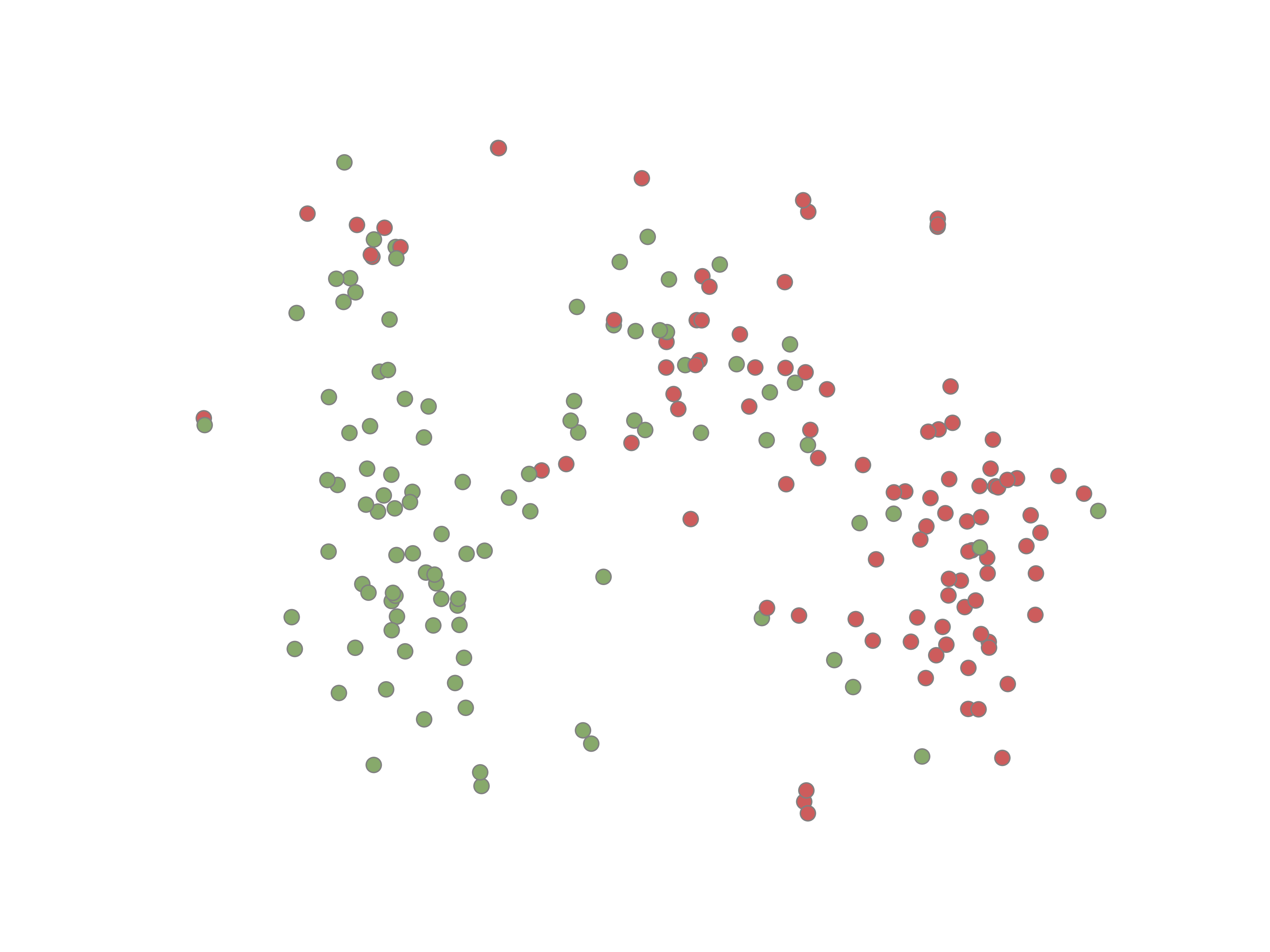}
    \includegraphics[width=0.3\linewidth]{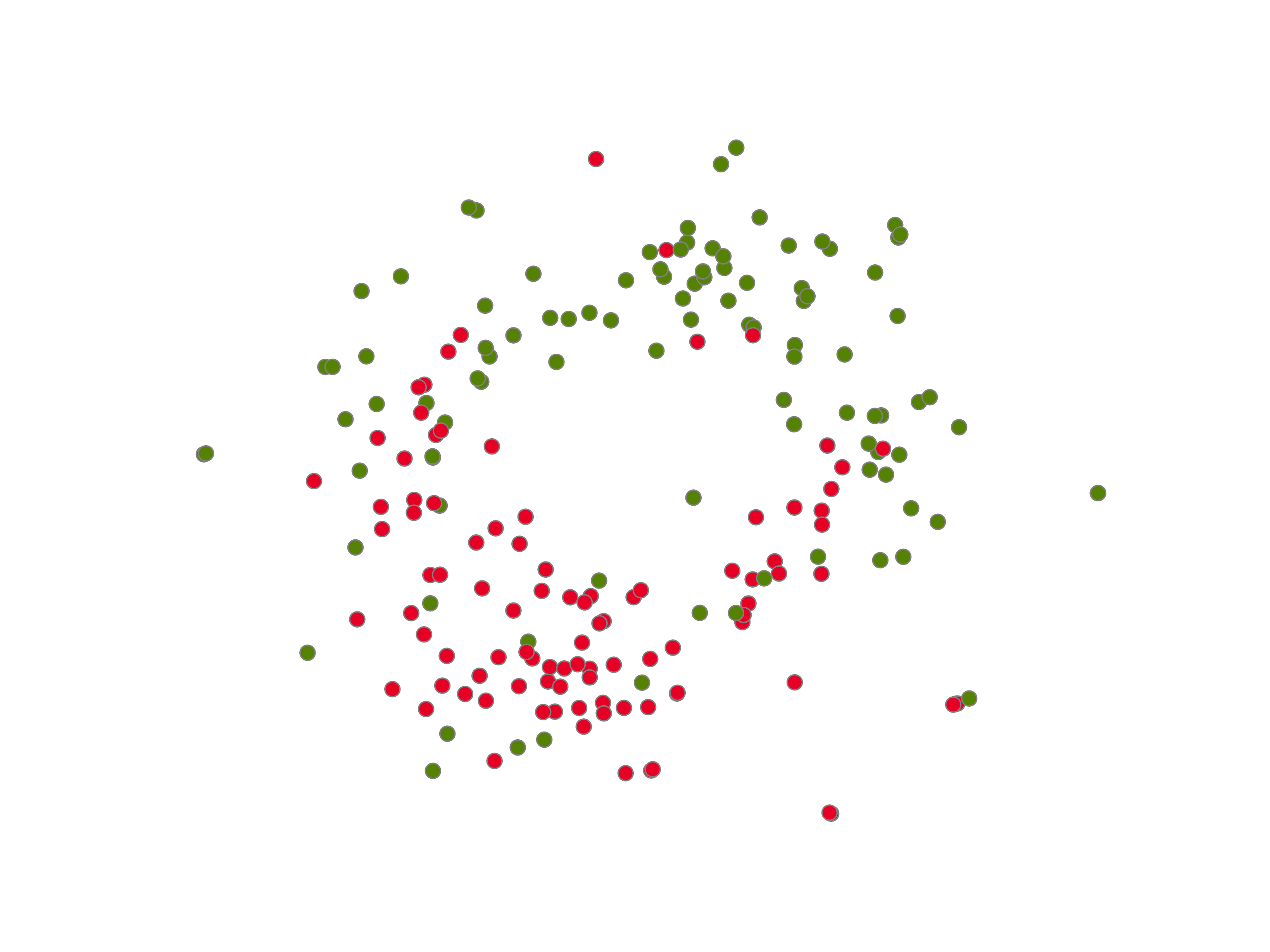}
    \includegraphics[width=0.3\linewidth]{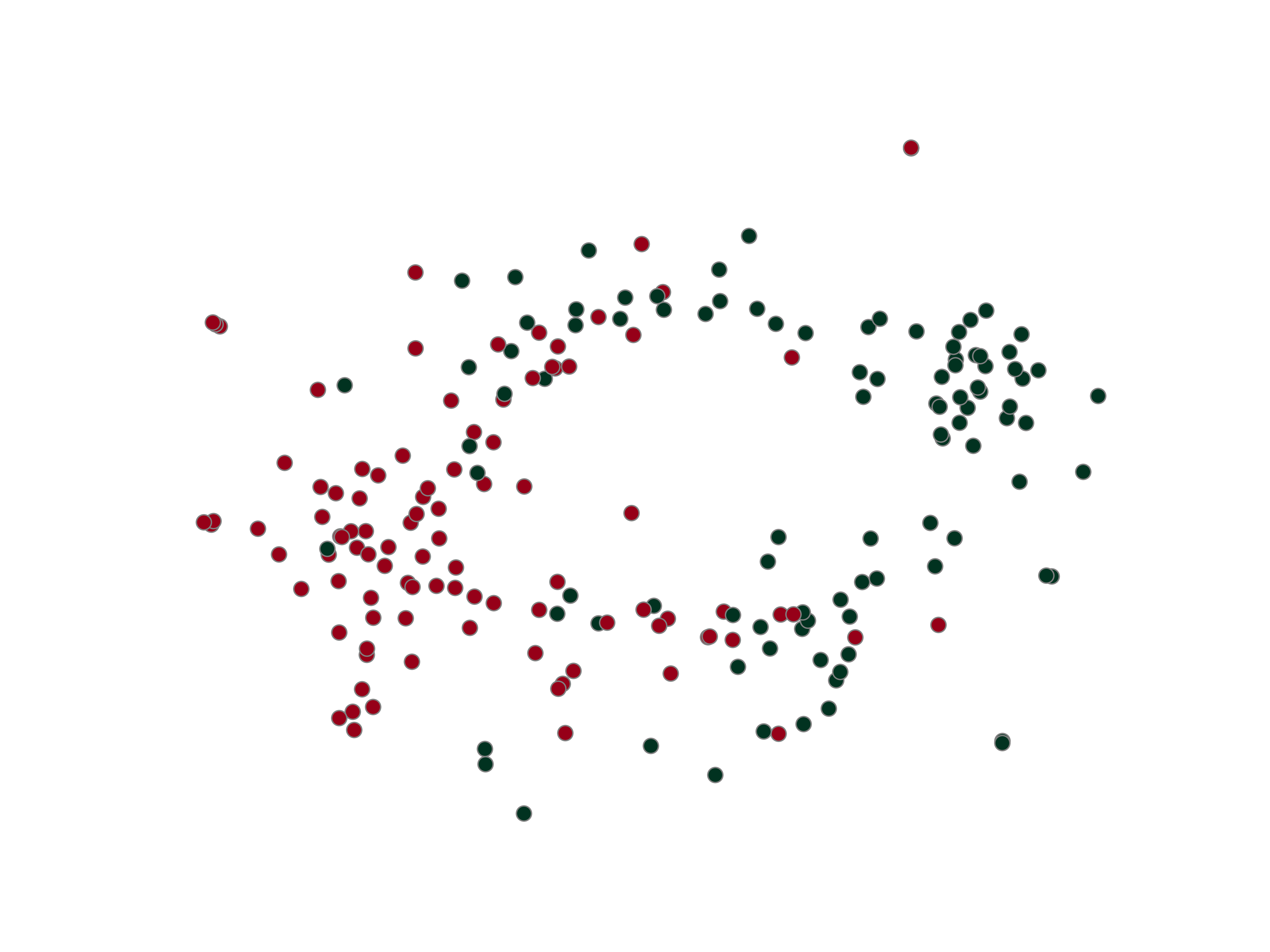}
    \includegraphics[width=0.3\linewidth]{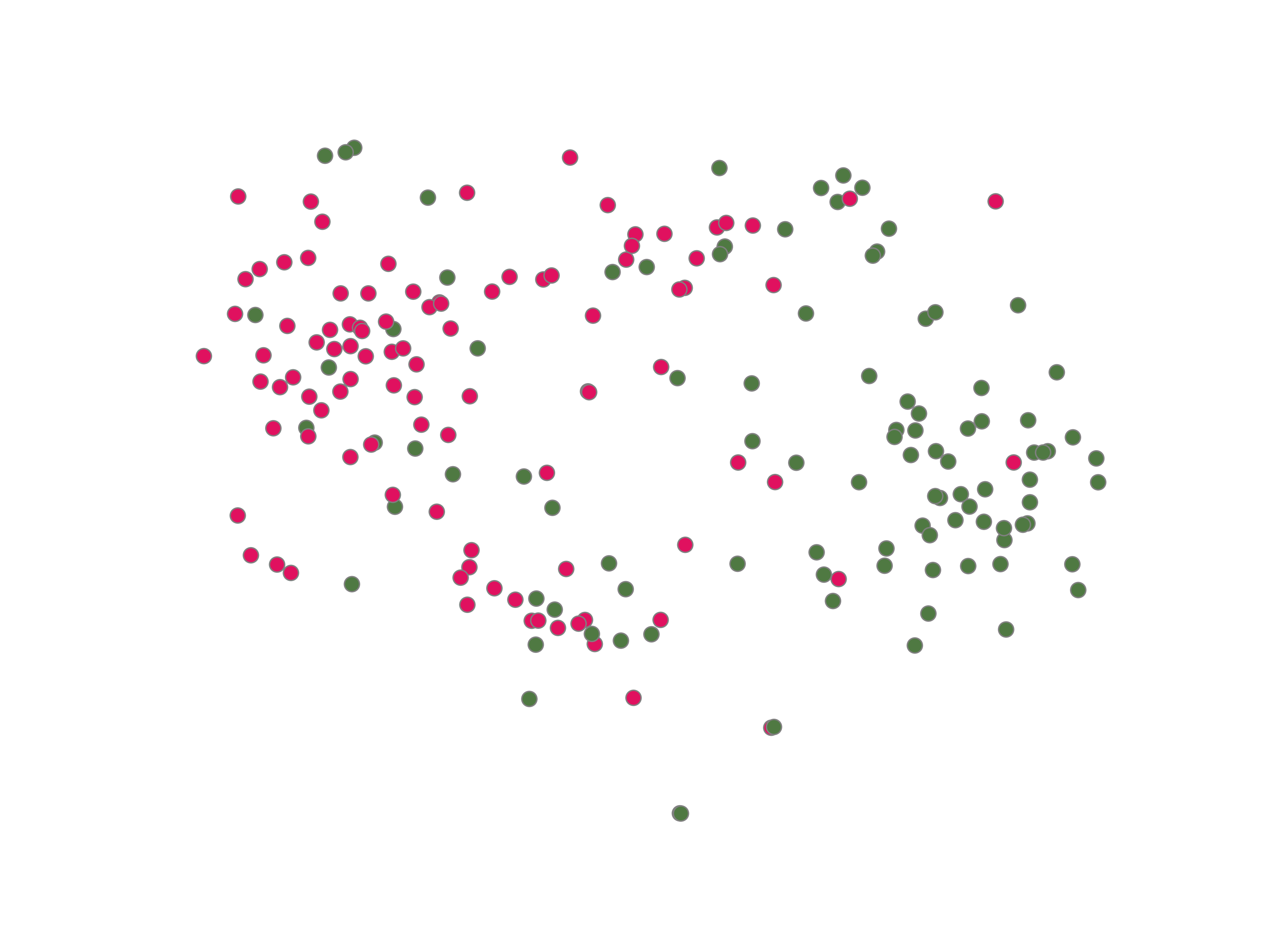}
    \includegraphics[width=0.3\linewidth]{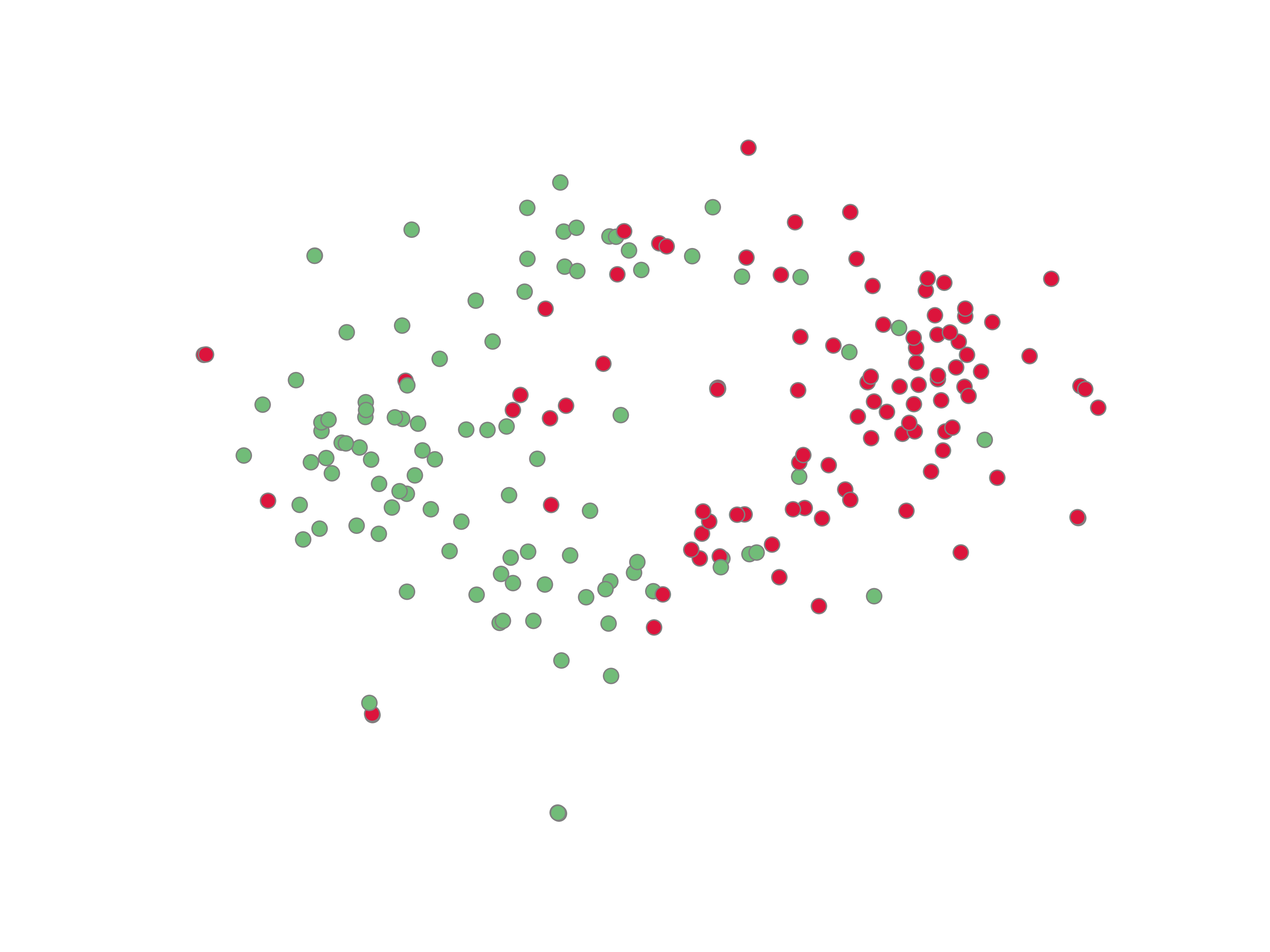}
    \includegraphics[width=0.3\linewidth]{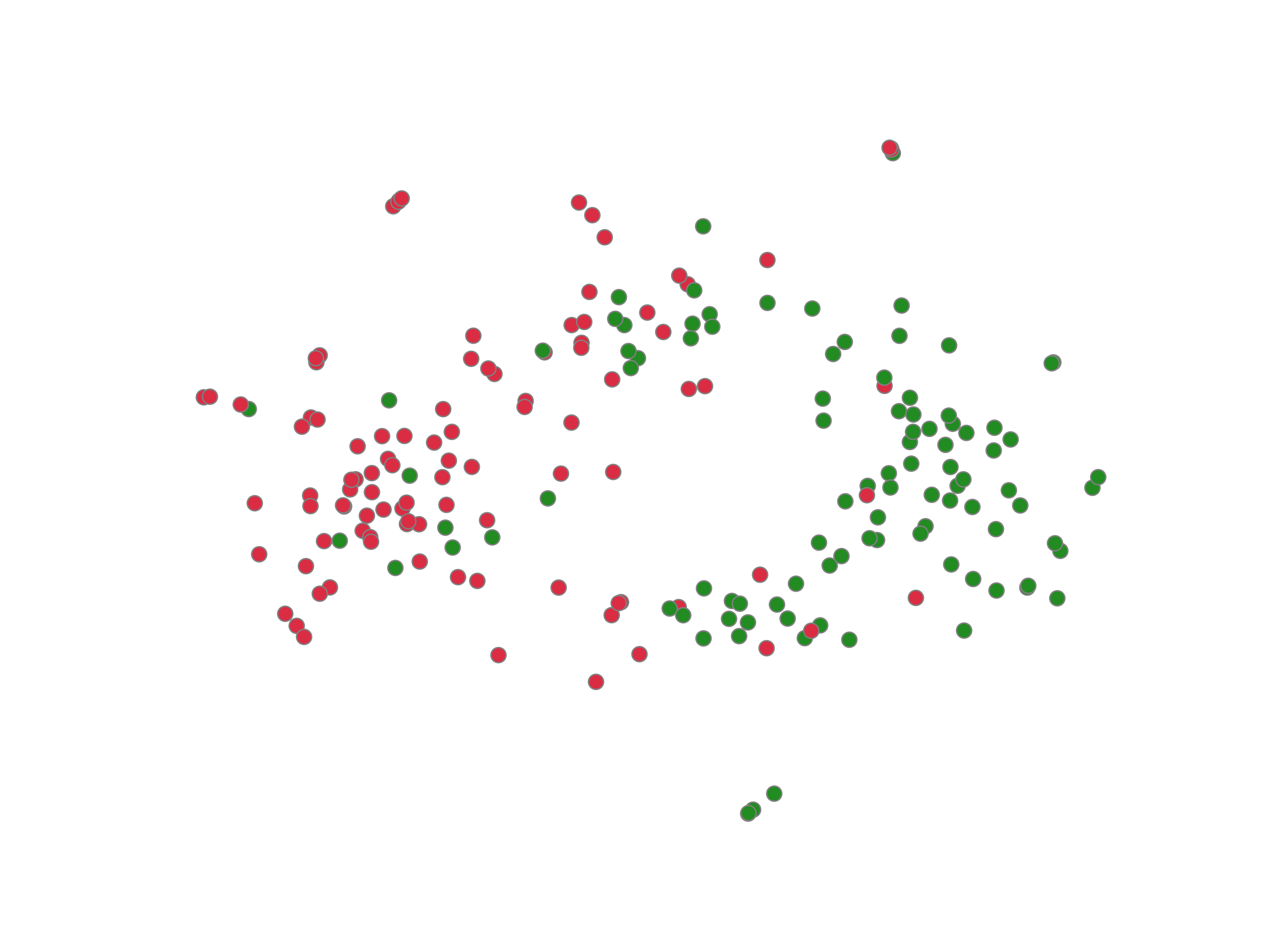}
    \includegraphics[width=0.3\linewidth]{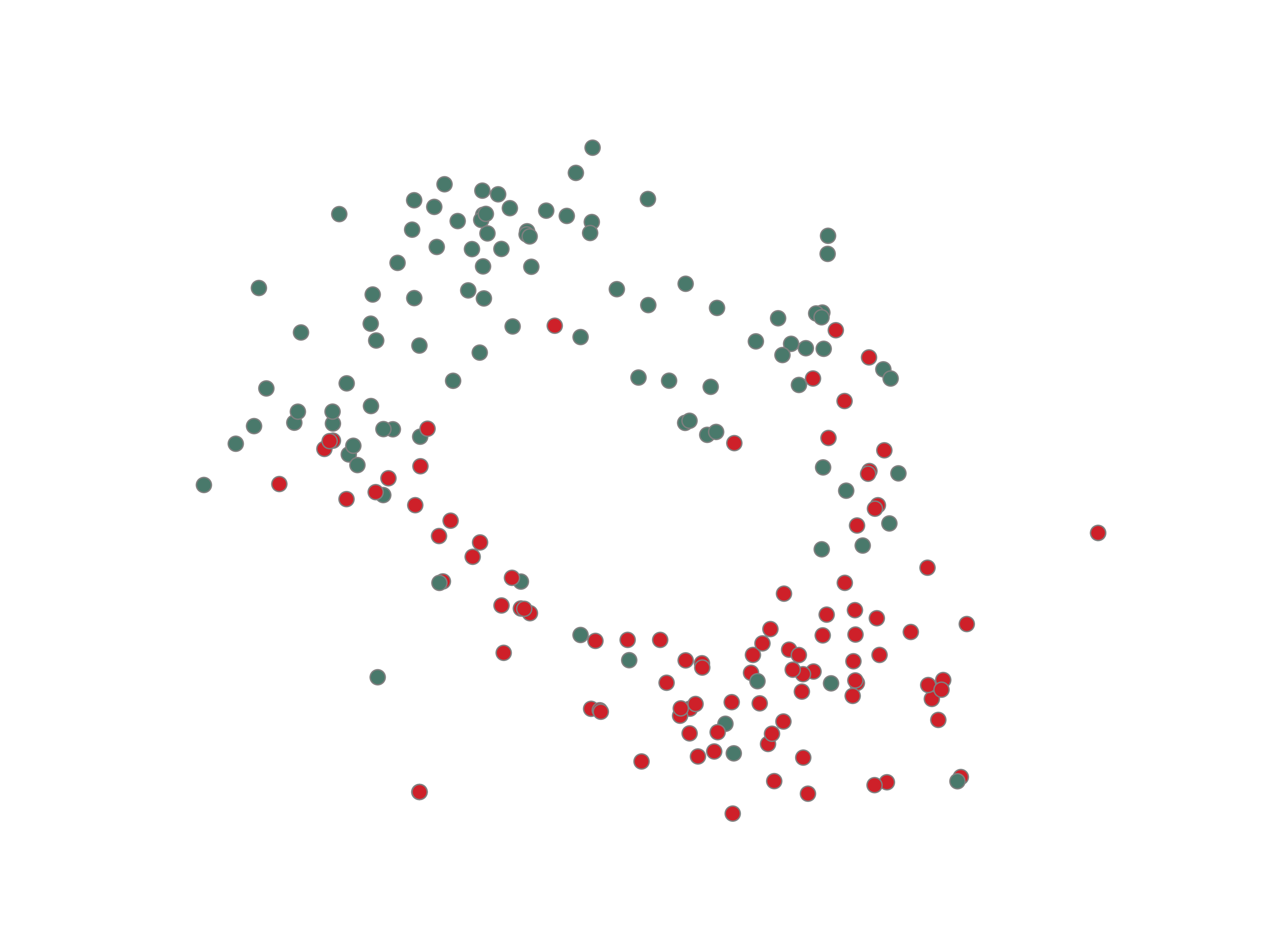}
    \includegraphics[width=0.3\linewidth]{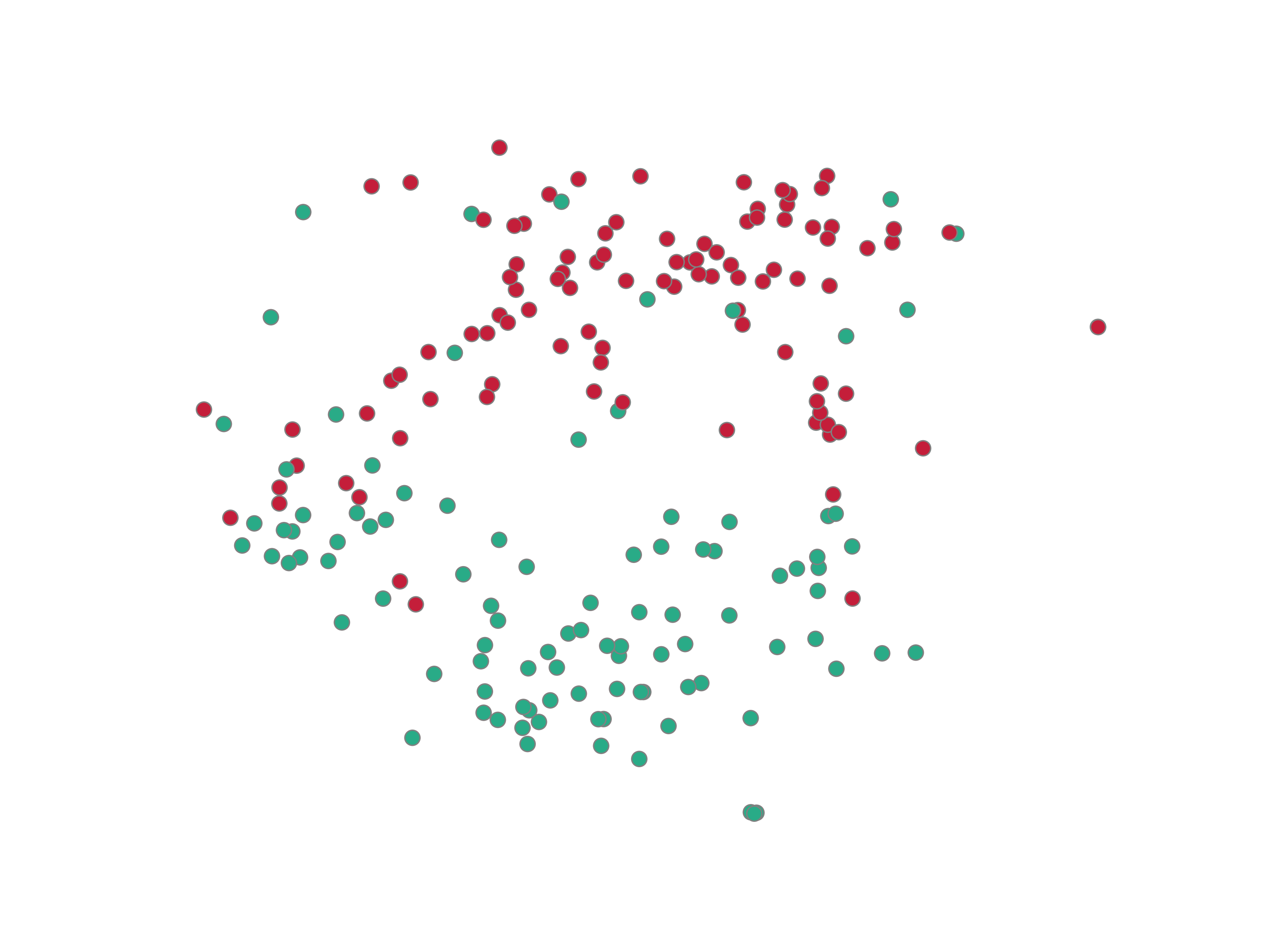}
    \caption{t-SNE projections of 200 learnt drug-drug representations, across 9 side-effects. Each side effect is colour-coded with a shade of red for positive drug-drug pairs and a shade of green for negative pairs.}
    \label{fig:tsne2}
\end{figure}

\begin{figure}
    \centering
    \includegraphics[width=\linewidth]{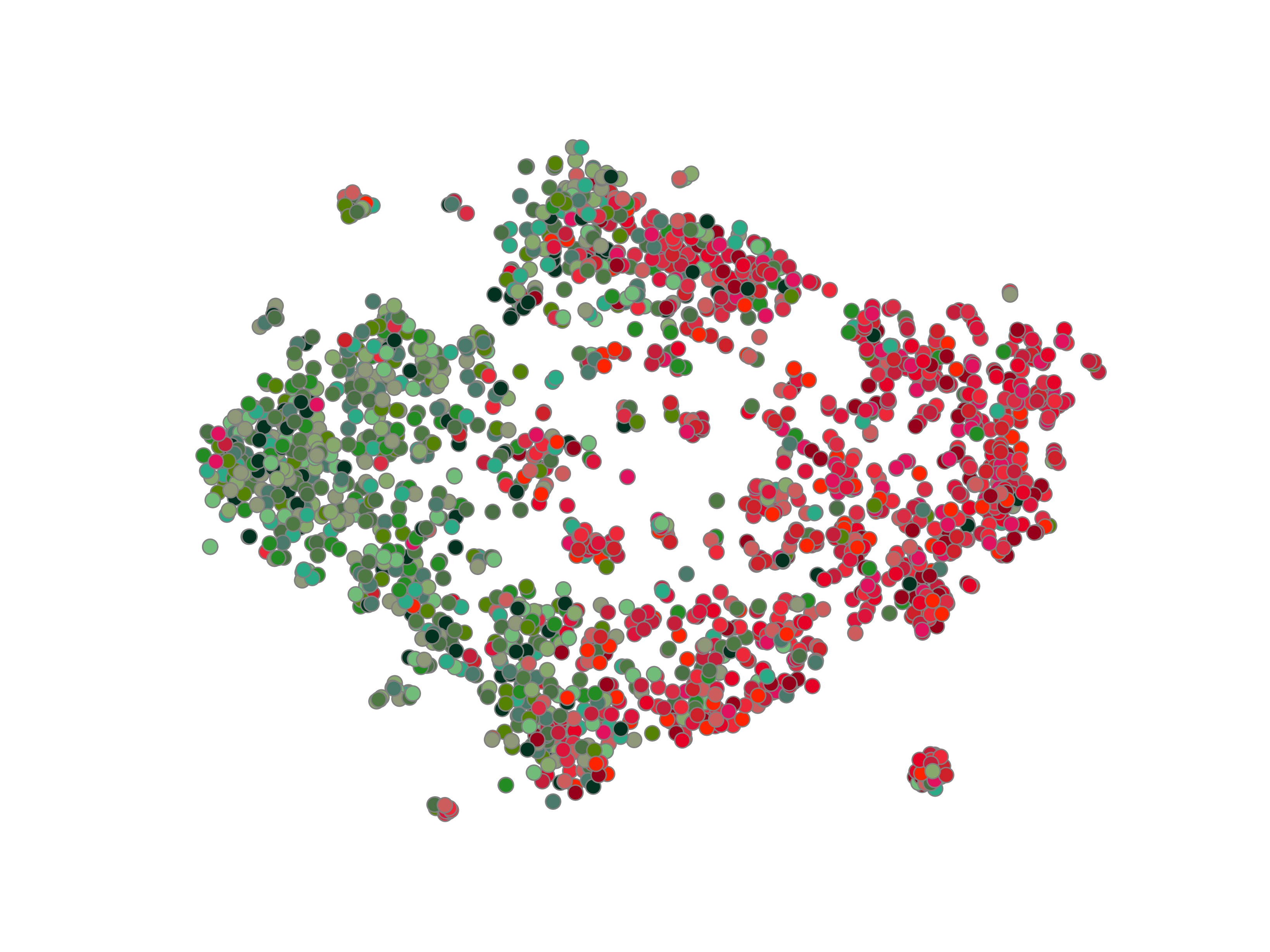}
    \caption{t-SNE projection of 2000 learnt drug-drug representations, sampled to cover 10 side effects. Each side effect is colour-coded with a shade of red for positive drug-drug pairs and a shade of green for negative drug-drug pairs.}
    \label{fig:tsne}
\end{figure}

The effectiveness of the learnt joint representations may be investigated \emph{qualitatively} as well, and to that end, we have devised a controlled visualisation experiment. Our objective is to investigate the distribution of learnt drug-drug embeddings with respect to individual side effects.\\ \\
We start with a pre-trained MHCADDI model. For each side effect in a sample of 10, we have randomly sampled 100 drug-drug pairs exhibiting this side effect, and 100 drug-drug pairs not exhibiting it (both unseen by the model during training). For these pairs, we derived their embeddings, and projected them into two dimensions using t-SNE \cite{tSNE}.\\ \\
The visualised embeddings may be seen in Figure \ref{fig:tsne2} for individual side-effects, and Figure \ref{fig:tsne} for a combined plot. These plots demonstrate that, across a single side-effect, there is a discernible clustering in the projected 2D space of drug-drug interactions. Furthermore, the combined plot demonstrates that strongly side-effect inducing drug-drug pairs tend to be clustered together, and away from the pairs that do not induce side-effects.

\section{Conclusions}
In this work, we have presented a graph neural network architecture which sets state-of-the-art results for predicting the possible polypharmacy side effects of drug combinations, using \emph{solely} the molecular structure information of drug pairs. By performing message passing within each drug, as well as co-attending over the other drug's structure, we demonstrated the power of integrating \emph{joint} drug-drug information during the representation learning phase for individual drugs. Future directions could include applying such cross-modal architectures to predicting interactions in different kinds of networks (such as language or social) where the components of different networks interact with each other.

\bibliography{references}
\bibliographystyle{ACM-Reference-Format}

\appendix

\newpage

\section{Hyperparameters}

In the interests of reproducibility, in this supplementary material we detail all the hyperparameters used for training and evaluating our MHCADDI model described above.\\ \\
For each atom, its features $a_i^{(d_x)}$ are obtained by concatenating the specified input features with a learnable embedding (unique to each atom) of 32 dimensions.\\ \\
The input node projection function, $f_i$, is a single-layer MLP with no bias:
\begin{equation}
    f_i(x) = {\bf W}_{f_i}x
\end{equation}
computing 32 features, followed by dropout with $p=0.2$.\\ \\
The learnable edge embeddings, $e_{ij}^{(d_x)}$, are of 32 features, and dropout with $p=0.2$ is immediately applied to them once retrieved.\\ \\
The node projection MLP for message passing, $f_v^t$, is a single-layer MLP with no bias:
\begin{equation}
    f_v^t(x) = {\bf W}_{f_v^t}x
\end{equation}
computing 32 features, followed by dropout with $p=0.2$.\\ \\
The edge projection MLP for message passing, $f_e^t$, is a two-layer MLP with leaky ReLU \cite{LeakyReLU} activations:
\begin{equation}
    f_e^t(x) = \text{LeakyReLU}\left({\bf W}_{f_e^t}^2 \text{LeakyReLU}\left({\bf W}_{f_e^t}^1x + b_{f_e^t}^1\right) + b_{f_e^t}^2\right)
\end{equation}
Both layers compute 32 features, and dropout with $p=0.2$ is applied to both of their outputs.\\ \\
The key/value projection matrices, ${\bf W}_k^t$ and ${\bf W}_v^t$, of each attention head, compute 32 features each.\\ \\
The outer message MLP for co-attention, $f_o^t$, is a single-layer MLP with the leaky ReLU activation:
\begin{equation}
    f_o^t(x) = \text{LeakyReLU}\left({\bf W}_{f_o^t}x + b_{f_o^t}\right)
\end{equation}
computing 32 features, followed by dropout with $p=0.2$.\\ \\
The readout projection MLP, $f_r$, is a single-layer MLP with the leaky ReLU activation:
\begin{equation}
    f_r(x) = \text{LeakyReLU}\left({\bf W}_{f_r}x + b_{f_r}\right)
\end{equation}
The head node and tail node mapping matrices, ${\bf M}_h$ and ${\bf M}_t$ (used within the computation of side effect likelihoods), compute 32 features each.\\ \\
The margin hyperparameter $\gamma$ has been set to $\gamma = 1$.\\ \\

\end{document}